\documentclass{article}

\usepackage{microtype}
\usepackage{graphicx}
\usepackage{booktabs} 
\usepackage{tikz}
\usetikzlibrary{positioning, arrows.meta}
\usepackage{amsmath}
\usepackage{amssymb}
\usepackage{mathtools}
\usepackage{amsthm}
\usepackage{algorithmic}
\usepackage{textcomp}
\usepackage{xcolor}
\usepackage{multirow}
\usepackage{float}
\usepackage{lineno}
\usepackage{hyperref}

\usepackage{placeins}
\usepackage[accepted]{icml_template/icml2026}

\theoremstyle{plain}

\theoremstyle{definition}

\theoremstyle{remark}

\usepackage[textsize=tiny]{todonotes}

\icmltitlerunning{Evaluation of Contextual Understanding in Large Language Models}

\begin{document}

\twocolumn[
\icmltitle{Evaluation of Contextual Understanding in Large Language Models}

\icmlsetsymbol{equal}{*}

\begin{icmlauthorlist}
\icmlauthor{Subavarshana Arumugam}{equal,unimelb}
\icmlauthor{Mamta Nallaretnam}{equal,unimelb}
\icmlauthor{Kithuni Wickramasinghe}{equal,unimelb}
\icmlauthor{Chamath Gunapala}{equal,unimelb}
\icmlauthor{Pragatheeswaran Vipulanandan}{miami}
\icmlauthor{Uthayasanker Thayasivam}{unimelb}
\icmlauthor{Kamal Premaratne}{miami}
\end{icmlauthorlist}

\icmlaffiliation{unimelb}{Department of Computer Science and Engineering, University of Moratuwa, Moratuwa, Sri Lanka}
\icmlaffiliation{miami}{Department of Electrical and Computer Engineering, University of Miami, Coral Gables, Florida, USA}

\icmlcorrespondingauthor{Mamta Nallaretnam}{nallaretnam.21@cse.mrt.ac.lk}

\icmlkeywords{Machine Learning, ICML, Large Language Models, Knowledge Graphs, Contextual Understanding}

\vskip 0.3in
]

\printAffiliationsAndNotice{\icmlEqualContribution}


\begin{abstract}
Large Language Models (LLMs) demonstrate impressive performance across diverse NLP tasks, yet their ability to exhibit genuine contextual understanding remains uncertain. Traditional evaluation metrics such as perplexity, BiLingual Evaluation Understudy (BLEU), or surface-level accuracy fail to reveal how well LLMs extract, integrate, and reason over contextual information--a gap particularly critical in question answering, where models must align responses with contextually grounded knowledge rather than memorized associations. We propose a novel knowledge graph-based evaluation framework introducing Semantic Structural Similarity for KGs (S3KG), a hybrid similarity measure integrating structural and semantic similarity into a continuous evaluation score, alongside a diagnostic framework for categorizing reasoning errors. To validate this pipeline, we evaluate S3KG against established metrics on a curated question-answer (QA) benchmark, demonstrating its effectiveness in measuring correctness, faithfulness, and interpretability in LLM-generated responses.
\end{abstract}


\section{Introduction}


Large language models (LLMs) have transformed natural language processing, demonstrating strong performance on tasks ranging from open-domain question answering (QAing) to complex reasoning \cite{izacard2021fid, wei2022cot}. Yet a fundamental question persists: do these models genuinely understand the context they process, or do they exploit statistical correlations to produce plausible outputs without true comprehension? This is especially consequential in high-stakes domains such as medical and legal QAing, where responses must be grounded in the provided context rather than memorized priors.


\section{Related Work}
\label{sec:related}


\subsection{LLM Evaluation and Contextual Understanding}


Existing evaluation metrics--perplexity, BLEU~\cite{bleu}, and token-level accuracy--measure surface-level fluency and overlap but fail to capture relational depth or factual faithfulness. \citet{zhu2024can} benchmark LLMs across several structured tasks, including co-reference resolution and dialogue state tracking, finding that models capture general context patterns but fail in fine-grained interpretation. \citet{mesaqa} probe reasoning fidelity by manipulating in-context examples through logical substitutions, revealing systematic failures in formal reasoning that surface-level metrics do not expose. 

On the construction side, LLM-driven KG frameworks such as GraphRAG~\cite{edge2024} and KEA~\cite{kea} have demonstrated that high-quality relational triplets can be extracted with minimal supervision, while hallucination-oriented evaluators like GraphEval~\cite{grapheval} adopt few-shot prompting with instruction tuning to encourage label consistency across graphs. Despite these advances, no existing metric jointly captures relational structure and semantic faithfulness in a continuous  interpretable score.


\subsection{Graph Similarity Methods}


Embedding-based approaches such as TransE~\cite{bordes2013translating} and RotatE~\cite{sun2019rotate} learn entity representations over fixed vocabularies, making them ill-suited for cross-graph comparison where entity sets are disjoint. The WL graph kernel~\cite{shervashidze2011weisfeiler} offers training-free structural comparison but treats labels as opaque symbols, penalizing semantically equivalent but lexically distinct terms. The WWL kernel~\cite{togninalli2019wasserstein} partially addresses this with Wasserstein node-distribution comparison, yet still lacks semantic label grounding. KEA~\cite{kea} combines WL kernels with SBERT~\cite{reimers2019sbert} clustering to bridge the lexical gap, but its cluster-merge is lossy because merging semantically close labels discards fine-grained relational distinctions. Taken together, these methods either treat labels as opaque symbols or rely on lossy clustering, leaving a clear need for a similarity 
measure that preserves fine-grained relational distinctions while handling semantic variation.


\subsection{Our Contributions}


Our work makes three main contributions. 
\begin{itemize}
    \item \textbf{Semantic structural similarity for KGs (S3KG)} is a hybrid structural--semantic similarity metric that converts LLM responses and reference answers into knowledge graph (KG) triplets and produces a continuous interpretable evaluation score.
    \item The \textbf{Contextual Understanding Score (CUS)} is a model-level aggregate of two complementary dimensions: factual accuracy and contextual faithfulness, enabling cross-model comparison across benchmarks.
    \item The \textbf{Triplet Analyzing Unit (TAU)} is a diagnostic component that categorizes reasoning errors at the triplet level for fine-grained behavioral analysis.
\end{itemize}


\section{Methodology}
\label{sec:methodology}



\begin{figure}[H]
\centering
\includegraphics[width=0.95\linewidth]{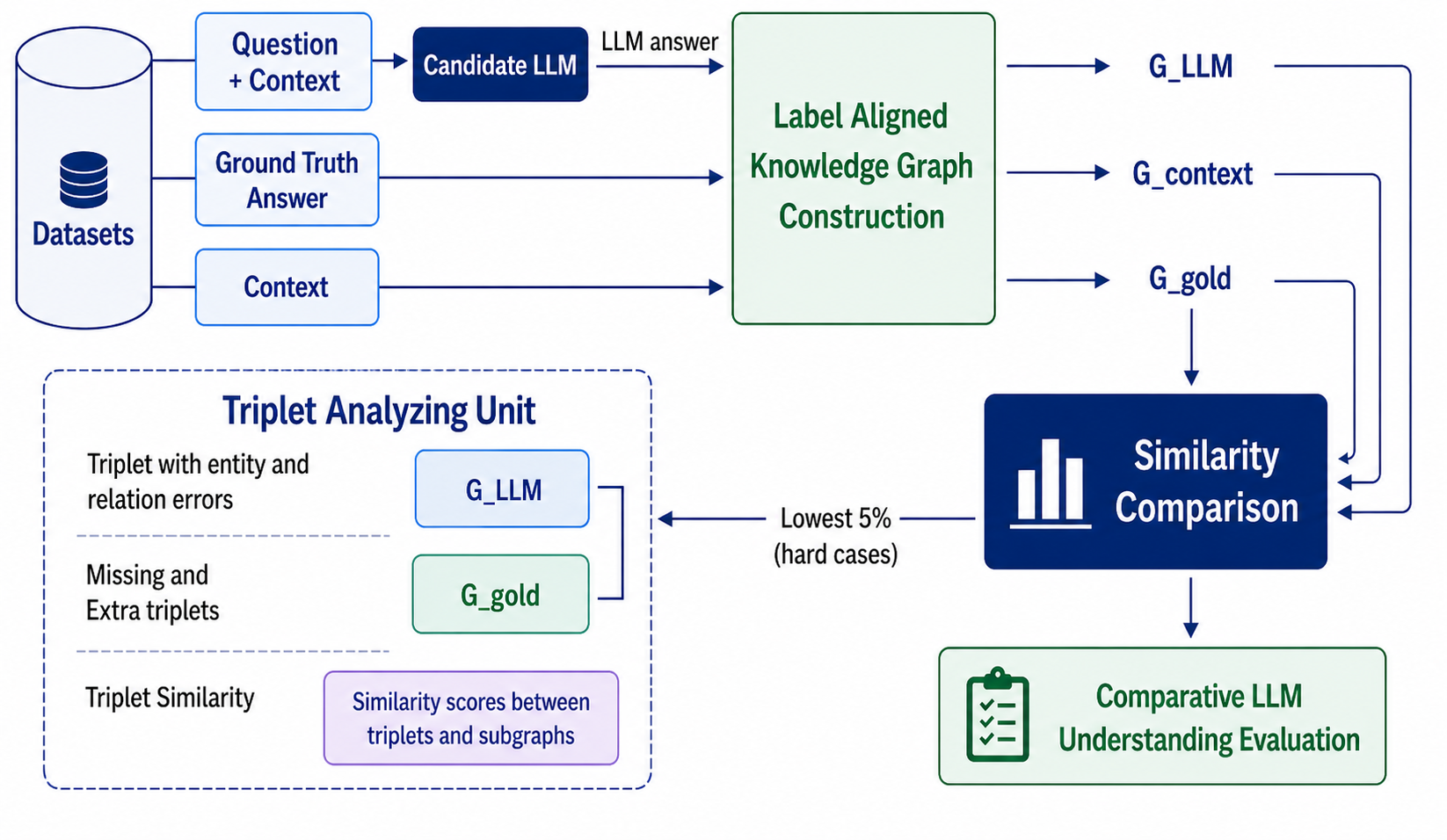}
\caption{KG-based evaluation pipeline. Given a QA pair, three KGs are constructed from the LLM response, gold answer, and supporting context, then compared via S3KG to produce GoldSim, CtxSim, and CUS.}
\label{fig:method}
\end{figure}


As illustrated in Fig.\ref{fig:method}, initially the candidate LLM is prompted with the question alongside its relevant context. The generated response is collected as the LLM answer, which together with the ground truth answer and the supporting context is used for the knowledge graph construction.


\subsection{Knowledge Graph Construction}


Three KGs are constructed per QA instance from the gold answer, model-generated response, and supporting context, using the few-shot prompting strategy with instruction tuning from~\citet{grapheval} and \citet{kea} applied uniformly across all three sources to ensure a consistent label space. Following extraction, all entity and relation labels are normalized via lowercasing and lemmatization to eliminate residual surface-level variation, ensuring the three graphs are structurally compatible for meaningful comparison.


\subsection{S3KG: Semantic Structural Similarity for KGs}
\label{sec:S3KG}


S3KG computes similarity between two graphs $G_1$ and $G_2$ at two levels: at the \textit{triplet level,} individual facts are matched for fine-grained correspondence; at the \textit{graph level,} overall topology is compared. Both structural and semantic similarity are considered at each level. The full pipeline is illustrated in Appendix~\ref{sec:appendix_s3kg} (Figure~\ref{fig:s3kg_pipeline}).


\textbf{Triplet-Level Matching.}
Each triplet $(h, r, t)$ is serialized into a natural language (NL) string and encoded with SBERT~\cite{reimers2019sbert} (\texttt{paraphrase-MPNet-base-v2}). For each triplet in $\mathcal{T}_1$, the most semantically similar triplet in $\mathcal{T}_2$ is identified by cosine similarity, forming a filtered set $\widehat{\mathcal{T}}_2 \subseteq \mathcal{T}_2$ that anchors the structural comparison to semantically relevant content.


\textbf{Soft Label Alignment.}
Standard Weisfeiler–Lehman (WL) kernels treat lexically distinct but semantically equivalent labels as entirely disjoint. S3KG resolves this by independently aligning entity and relation labels: each label $\ell$ in $G_1$ is replaced by its closest counterpart in $G_2$ (by SBERT cosine similarity) whenever the similarity exceeds a threshold (we use $0.65$); otherwise it is left unchanged. Entity and relation labels are aligned separately to prevent cross-type collisions, yielding aligned graphs $\widetilde{G}_1$ and $\widetilde{G}_2$.


\textbf{Structural Similarity via WL Kernel.}
The normalized WL kernel~\cite{shervashidze2011weisfeiler} score, with multiple iterations (we use $5$) to capture multi-hop neighbourhood patterns over the aligned graphs) yields structural similarity as
\begin{equation}
  \text{S}_{\text{WL}}(\widetilde{G}_1,\,\widetilde{G}_2) 
    = \frac{%
      K(\widetilde{G}_1,\,\widetilde{G}_2)}{%
      \sqrt{K(\widetilde{G}_1,\, \widetilde{G}_1)
      \cdot 
      K(\widetilde{G}_2,\, \widetilde{G}_2)}}.
\label{eq:wl-score}
\end{equation}


\textbf{Semantic Similarity via SBERT Mean-Pool.}
Each graph is represented by the mean SBERT embedding of its triples. The semantic similarity $\text{S}_{\text{SBERT}}(\mathcal{T}_1,\widehat{\mathcal{T}}_2)$ is the cosine between these pooled representations, clipped to $[0,1]$ to discard negative correlations.


\textbf{Final Score.}
The structural and semantic scores are blended via mixing coefficient $\alpha \in [0,1]$ (we use $\alpha = 0.5$) as 
\begin{equation}
  \text{S}_{\text{S3KG}} 
    = (1-\alpha)\, \text{S}_{\text{WL}} + \alpha\, \text{S}_{\text{SBERT}}.
\label{eq:S3KG}
\end{equation}



\subsection{Triplet Analyzing Unit (TAU)}


For the 5\% of lowest-similarity QA pairs, a triplet analysis unit is applied to examine how the LLM-generated KG diverges from the ground truth. Each triplet is converted into an NL sentence and encoded using a Sentence Transformer, and cosine similarity is computed between ground truth and LLM triplets, with scores above a threshold (we use $0.76$) treated as aligned. After removing aligned triplets, the remaining pairs are categorized into four error types: relation wrong (entities match but relation differs), entity wrong (relation aligns but entities differ), extra triplets (hallucinated by the LLM), and missing triplets (not captured by the LLM).


\subsection{Datasets and Models}


Two datasets are employed in this study for their long-form answer coverage: PubMedQA~\cite{jin2019pubmedqa}, comprising biomedical QA pairs drawn from research articles, and MesaQA~\cite{wang-etal-2025-mesaqa}, comprising consumer healthcare QA pairs requiring multi-span evidence integration. Three instruction-tuned 7B-parameter models are evaluated: \texttt{Llama-2-7b-chat-hf}, \texttt{Gemma-7b-it}, and \texttt{Mistral-7B-Instruct-v0.2}, with \texttt{Falcon-7B} included as a baseline. Responses are collected across a temperature sweep of $\{0.0, 0.3, 0.7, 1.0\}$ to analyze the effect of generation stochasticity on contextual faithfulness. Full temperature sensitivity results appear in Appendix~\ref{sec:appendix_temp}.

\section{Experiments and Results}
\label{sec:results}


Our evaluation targets the full KG-based LLM evaluation pipeline (Fig.~\ref{fig:method}), which comprises two core components: 
    (1)~\textbf{KG construction}, which extracts structured representations from the LLM response, gold answer, and supporting context; and
    (2)~\textbf{S3KG similarity module}, which compares these graphs to produce the Comparative LLM Understanding Score (CUS). To rigorously assess S3KG's similarity scoring in isolation, we first benchmark it independently across nine semantic equivalence datasets in \S\ref{sec:benchmark}, before evaluating the complete pipeline on QA benchmarks in \S\ref{sec:llm_eval}.


\subsection{S3KG Benchmark Evaluation}
\label{sec:benchmark}


\textbf{Datasets and Task Formulation.}
We evaluate on nine datasets spanning three structural categories, each cast as a binary semantic equivalence task ($N{=}400$, balanced), with performance measured by maximum F1 via threshold sweep. The \textbf{short-text} category comprises MRPC~\cite{dolan2005mrpc}, PAWS-Wiki~\cite{zhang2019paws}, and STS12~\cite{agirre2012semeval} (10--22 words). The \textbf{KG-perturbed paragraph} category comprises five datasets (SK-Codex~400, SK-Combined, SK-FindKG, SK-GloBI, SK-Oregano) built by perturbing entity relationships in KG-derived paragraphs (69--126 words) across encyclopedic~\cite{safavi2021codex}, financial~\cite{findkg}, biological~\cite{globiref}, and food ontology~\cite{oregano} domains. The \textbf{Wikipedia Entity-Swap} dataset (399 pairs) serves as an anti-circularity control, constructed via NLP-based perturbations with no KG involvement~\cite{sennrich2016neural, wei2019eda}. Baselines include ROUGE-1/2/L~\cite{lin2004rouge}, BLEU~\cite{bleu}, BERTScore~\cite{zhang2019bertscore}, MiniLM~\cite{wang2020minilm}, and Sentence-T5-base~\cite{ni2022t5}. S3KG is evaluated with $\alpha \in \{0.0, 0.1, \ldots, 1.0\}$; we report the best-performing $\alpha$ per dataset alongside AUROC, with the full sweep in Appendix~\ref{sec:appendix_alpha}.


\textbf{Results.}
Table~\ref{tab:combined} reports F1 and AUROC across all benchmarks. S3KG achieves top-1 or top-2 F1 on 7 of 9 datasets. On KG-perturbed paragraphs, structural signals are most valuable: S3KG gains up to $+7.6$ F1 points (SK-GloBI) and correctly captures relational role distinctions where ROUGE-1 collapses to near-random (PAWS-Wiki AUC $= 0.490$, S3KG F1 $= 0.766$). The exception is SK-FindKG, where financial vocabulary introduces KG extraction noise and Sentence-T5-base leads (F1 $= 0.848$), highlighting extraction quality as a bottleneck. On short texts, dense models dominate due to sparse relational structure (Sentence-T5-base: MRPC $= 0.766$, STS12 $= 0.853$). On the Wikipedia Entity-Swap control, S3KG achieves the strongest result (F1 $= 0.872$ vs MiniLM $0.821$), confirming gains are not an artifact of the KG pipeline. Optimal $\alpha$ varies by dataset and is further discussed in Appendix~\ref{sec:appendix_alpha}.

\begin{table*}[t]
    \centering
    \caption{F1 / AUROC scores across all benchmark datasets. \textbf{Bold} indicates best value per column. S3KG uses the best-performing $\alpha$ per dataset selected by grid search (full sweep in Appendix~\ref{sec:appendix_alpha}). $^\dagger$ROUGE-1 excluded from Wiki Swap due to surface-form artefact. Best $\alpha$: MRPC~0.3, PAWS~0.5, STS12~0.1, C400~0.5, Comb.~0.5, Find~0.0, GloBI~0.6, Oreg.~0.4, W-Swap~0.1.}
    \label{tab:combined}
    \small
    \setlength{\tabcolsep}{3pt}
    \renewcommand{\arraystretch}{1.08}
    \begin{tabular}{l ccc ccccc c}
        \toprule
        {}
            & \multicolumn{3}{c}{\textbf{Short Text}}
            & \multicolumn{5}{c}{\textbf{KG-Perturbed Paragraphs}}
            & \textbf{Anti-Circ.} \\
        \cmidrule(lr){2-4} \cmidrule(lr){5-9} \cmidrule(lr){10-10}
        \textbf{Method}
            & \textbf{MRPC} & \textbf{PAWS} & \textbf{STS12}
            & \textbf{C400} & \textbf{Comb.} & \textbf{Find} & \textbf{GloBI} & \textbf{Oreg.}
            & \textbf{W-Swap} \\
        \midrule
        S3KG (Ours)
            & 0.692/0.673 & \textbf{0.766}/\textbf{0.795} & 0.786/0.834
            & \textbf{0.932}/\textbf{0.973} & \textbf{0.834}/\textbf{0.829} & 0.767/0.796 & \textbf{0.892}/\textbf{0.935} & \textbf{0.812}/\textbf{0.892}
            & \textbf{0.872}/\textbf{0.890} \\
        \midrule
        ROUGE-1
            & 0.745/0.784 & 0.678/0.490 & 0.725/0.754
            & 0.835/0.917 & 0.732/0.728 & 0.745/0.745 & 0.784/0.833 & 0.745/0.782
            & ---$^\dagger$ \\
        ROUGE-2
            & 0.720/0.721 & 0.715/0.721 & 0.681/0.656
            & 0.822/0.894 & 0.707/0.711 & 0.717/0.706 & 0.776/0.833 & 0.752/0.791
            & 0.860/0.772 \\
        ROUGE-L
            & 0.729/0.760 & 0.735/0.807 & 0.703/0.710
            & 0.792/0.855 & 0.722/0.717 & 0.719/0.721 & 0.763/0.800 & 0.792/0.835
            & 0.729/0.311 \\
        BLEU
            & 0.687/0.677 & 0.716/0.747 & 0.671/0.644
            & 0.806/0.884 & 0.715/0.708 & 0.711/0.710 & 0.775/0.819 & 0.745/0.794
            & 0.868/0.745 \\
        BERTScore
            & 0.758/0.816 & 0.691/0.702 & 0.682/0.636
            & 0.823/0.916 & 0.757/0.792 & 0.739/0.761 & 0.816/0.871 & 0.743/0.798
            & 0.747/0.645 \\
        MiniLM
            & 0.723/0.748 & 0.687/0.638 & 0.833/0.894
            & 0.875/0.943 & 0.770/0.789 & 0.802/0.844 & 0.780/0.817 & 0.773/0.814
            & 0.821/0.811 \\
        sent-T5
            & \textbf{0.766}/\textbf{0.816} & 0.674/0.668 & \textbf{0.853}/\textbf{0.928}
            & 0.876/0.944 & 0.770/0.827 & \textbf{0.848}/\textbf{0.902} & 0.728/0.760 & 0.797/0.871
            & 0.762/0.806 \\
        \bottomrule
        \multicolumn{10}{l}{%
        \footnotesize
        C400: SK-Codex~400;\enspace Comb.: SK-Combined;\enspace
        Find: SK-FindKG;\enspace Oreg.: SK-Oregano;\enspace
        W-Swap: Wikipedia Entity-Swap.}
    \end{tabular}
\end{table*}


\subsection{TAU Evaluation}


Table~\ref{tab:triplet} illustrates the TAU performace, on a manually annotated subset of the MesaQA and PubMed datasets, where aligned triplet pairs between ground-truth and LLM-generated KGs were labeled by three medical students, achieving strong agreement (pairwise F1: $0.97--0.99$). Our method uses sentence-level semantic similarity for alignment, while the KEA baseline relies on component-wise matching. Our approach improves Micro F1 ($0.895$ vs $0.836$) and Macro F1 ($0.782$ vs $0.622$), driven by a +34.8\% recall gain with a modest drop in precision. This enables robust alignment of semantically equivalent relations despite lexical variation, which are often missed by KEA. Overall, these results demonstrate that our method outperforms KEA in TAU  performance.

\begin{table}[H]
    \centering
    \caption{TAU performance vs.\ KEA baseline.}
    \label{tab:triplet}
    \begin{tabular}{l cc}
        \toprule
        \textbf{Metric} 
            & \textbf{KEA} & \textbf{Ours} \\
        \midrule
        Micro Precision 
            & \textbf{0.853} & 0.847 \\
        Micro Recall    
            & 0.738 & \textbf{0.949} \\
        Macro Precision 
            & \textbf{0.953} & 0.803 \\
        Macro Recall    
            & 0.641 & \textbf{0.946} \\
        Micro F1        
            & 0.836 & \textbf{0.895} \\
        Macro F1        
            & 0.622 & \textbf{0.782} \\
        \bottomrule
    \end{tabular}
\end{table}


\subsection{LLM Contextual Understanding Evaluation}
\label{sec:llm_eval}


To evaluate LLMs, S3KG is applied across three KGs per QA instance $i$: $\mathit{KG}_{\text{LLM}}^{(i)}$, $\mathit{KG}_{\text{gold}}^{(i)}$, and $\mathit{KG}_{\text{ctx}}^{(i)}$. \textbf{GoldSim}$(i) = \text{S}_{\text{S3KG}}(\mathit{KG}_{\text{LLM}}^{(i)}, \mathit{KG}_{\text{gold}}^{(i)})$ measures factual accuracy; \textbf{CtxSim}$(i) = \text{S}_{\text{S3KG}}(\mathit{KG}_{\text{LLM}}^{(i)}, \mathit{KG}_{\text{ctx}}^{(i)})$ measures contextual faithfulness. Since neither alone reflects true understanding, \textbf{CUS}  penalises imbalance between the two as 
\begin{equation}
  \text{CUS}(i) 
  = \frac{%
    2 
    \cdot 
    \text{GoldSim}(i) 
    \cdot \text{CtxSim}(i)}{%
    \text{GoldSim}(i) + \text{CtxSim}(i)}.
  \label{eq:cus_sample}
\end{equation}
The dataset-level score is the mean of $\text{CUS}(i)$ over all $N$ samples. Table~\ref{tab:llm} reports mean GoldSim, CtxSim, and CUS over $N = 400$ samples per model--dataset combination at $\alpha = 0.5$. Mistral-7B leads on CUS across both datasets (MesaQA: $0.678$; PubMedQA: $0.592$), driven by the highest CtxSim in each case ($0.736$ and $0.733$), reflecting strong contextual faithfulness. CtxSim exceeds GoldSim in all eight model--dataset combinations, indicating that instruction-tuned models systematically elaborate on context rather than producing concise reference-style responses. A consistent $\approx$10 percentage-point CUS gap between MesaQA and PubMedQA across all models reveals domain complexity, biomedical vocabulary, and reasoning as key bottlenecks rather than model size or architecture. Falcon-7B shows the weakest performance on both datasets and across all metrics.

On the 5\% lowest similarity pairs, Mistral-7B remains the most reliable, recovering more reference-aligned triplets than other models, while Gemma-7B is the weakest, particularly on MesaQA. PubMedQA hard cases show sharply lower aligned-triplet recovery across all models, confirming that biomedical terminology and entity variability dominate failure modes.

\begin{table}[t]
    \centering
    \caption{LLM evaluation results (mean over $N{=}400$ samples, $\alpha{=}0.5$).}
    \label{tab:llm}
    \setlength{\tabcolsep}{4pt}
    \begin{tabular}{l l ccc}
        \toprule
        \textbf{Dataset} 
            & \textbf{Model} 
            & \textbf{GoldSim} & \textbf{CtxSim} & \textbf{CUS} \\
        \midrule
        \multirow{4}{*}{MesaQA}
        {}
            & Gemma-7B        
            & \textbf{0.6889} & 0.7031 & 0.6774 \\
        {}
            & Llama-2-7B      
            & 0.6570 & 0.7236 & 0.6752 \\
        {}
            & Mistral-7B      
            & 0.6491 & \textbf{0.7356} & \textbf{0.6780} \\
        {}
            & Falcon-7B       
            & 0.6036 & 0.6458 & 0.6063 \\
        \midrule
        \multirow{4}{*}{PubMedQA}
        {}
            & Gemma-7B        
            & \textbf{0.5235} & 0.6401 & 0.5587 \\
        {}
            & Llama-2-7B      
            & 0.5220 & 0.6567 & 0.5651 \\
        {}
            & Mistral-7B      
            & 0.5138 & \textbf{0.7331} & \textbf{0.5923} \\
        {}
            & Falcon-7B       
            & 0.4541 & 0.5560 & 0.4800 \\
        \bottomrule
    \end{tabular}
\end{table}


\section{Discussion}
\label{sec:discussion}


S3KG moves beyond surface-level metrics via type-separated, one-to-one label alignment and WL kernel multi-hop sensitivity. The GoldSim/CtxSim decomposition exposes model-specific trade-offs invisible to aggregate metrics; the consistent PubMedQA--MesaQA domain gap confirms Mistral-7B handles domain-specific health knowledge more reliably than general health QA. KG extraction quality remains the primary bottleneck, and future work will incorporate GPT-4 and attention-based directional alignment to better distinguish semantically inverse relations.


\section{Conclusion}
\label{sec:con}


We presented a KG-based evaluation pipeline combining S3KG and the TAU for reproducible, interpretable measurement of LLM contextual understanding in QA. S3KG achieves best-per-dataset F1 of 0.766--0.932 and AUROC up to 0.973, consistently outperforming lexical and neural baselines on KG-rich datasets while remaining competitive on short-text settings. The framework is dataset-agnostic and readily extensible, supporting broader efforts toward trustworthy and verifiable AI evaluation.


\section*{Acknowledgments}

The work of Kamal Premaratne (KP) was supported by the U.S. National Science Foundation (NSF) under Award No. 2530256. The authors also acknowledge the developers and open-source communities behind the Falcon, Mistral, Llama, and Gemma large language models, as well as the Hugging Face platform for providing access to open-source models and tools that supported this research.

\bibliography{references}
\bibliographystyle{icml_template/icml2026}

\newpage
\appendix
\onecolumn

%
%
\section{S3KG Similarity Pipeline}
\label{sec:appendix_s3kg}


\begin{figure}[htbp]
\centering
\includegraphics[width=0.82\linewidth]{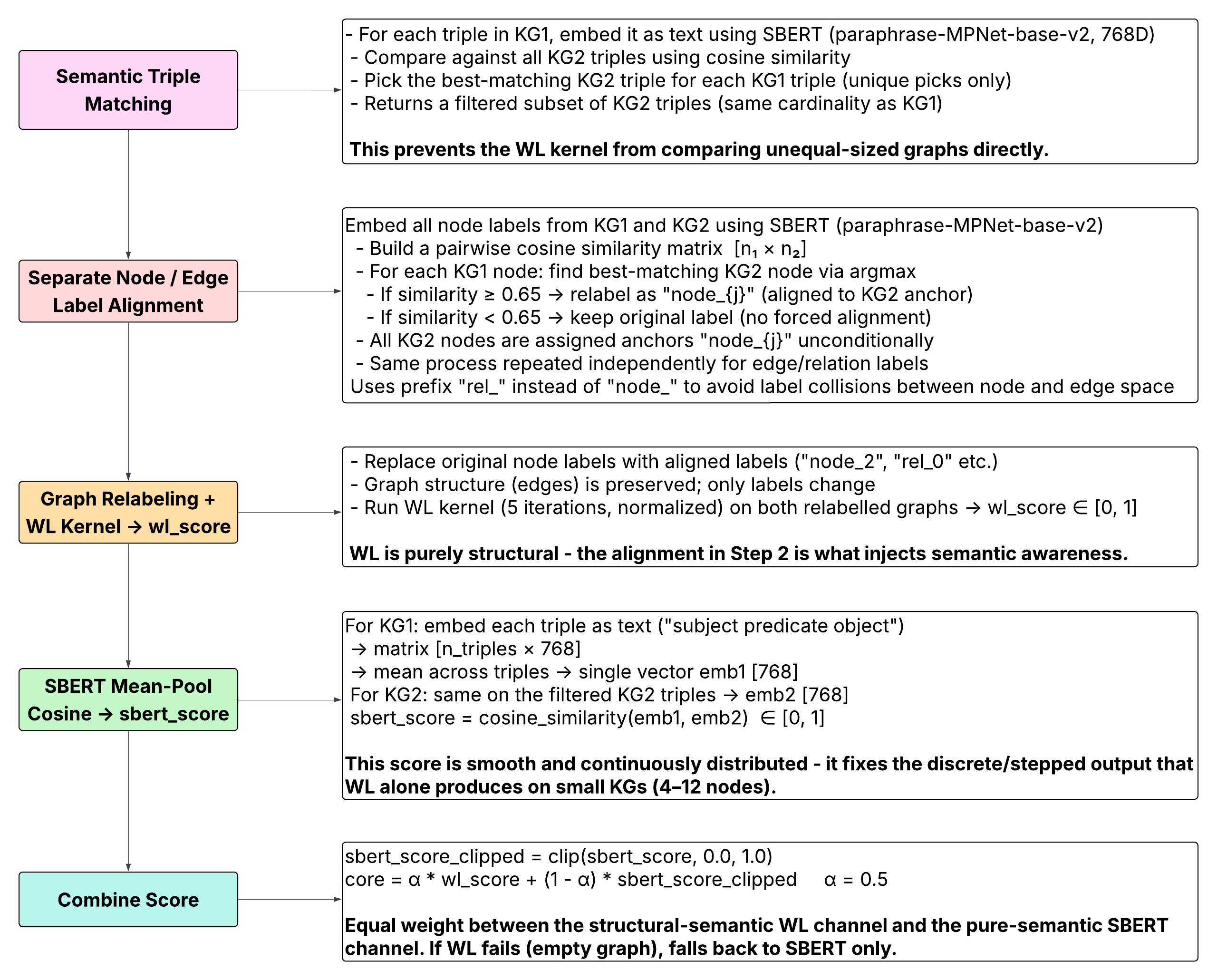}
\caption{S3KG similarity pipeline. Triple-level SBERT matching identifies semantically relevant triples in $G_2$; soft label alignment resolves lexical mismatches between entity and relation labels; the normalised WL kernel and SBERT mean-pool scores are blended via mixing coefficient $\alpha$ to produce the final S3KG score.}
\label{fig:s3kg_pipeline}
\end{figure}

%
%
\section{Hyperparameter $\alpha$ selection}
\label{sec:appendix_alpha}


This appendix reports the complete S3KG $\alpha$ sweep results across all evaluation datasets. Each table shows F1 and AUROC for $\alpha \in \{0.0, 0.1, \ldots, 1.0\}$,
where $\alpha = 0.0$ recovers a pure KG structural embedding and $\alpha = 1.0$ recovers a pure dense sentence-transformer representation, as defined in
Equation~\ref{eq:S3KG}. The best-performing $\alpha$ per dataset (by F1) is shown in \textbf{bold}.


\subsection*{B.1 Short-Text Datasets}


Table~\ref{tab:alpha_short} presents the $\alpha$ sweep for MRPC, PAWS-Wiki, and STS12. The optimal $\alpha$ differs notably across datasets: MRPC peaks at
$\alpha = 0.3$, PAWS-Wiki at $\alpha = 0.5$, and STS12 at $\alpha = 0.1$. The consistently low optimal values indicate that retaining a meaningful KG structural
component is beneficial for short-text paraphrase detection, and that a pure dense-embedding representation ($\alpha = 1.0$) is suboptimal for all three tasks.

\begin{table}[H]
    \centering
    \caption{S3KG $\alpha$ sweep on short-text datasets (F1 / AUROC).
    The best $\alpha$ per dataset by F1 is shown in \textbf{bold}.}
    \label{tab:alpha_short}
    \begin{tabular}{c cc cc cc}
        \toprule
        {}
            & \multicolumn{2}{c}{MRPC}
            & \multicolumn{2}{c}{PAWS-Wiki}
            & \multicolumn{2}{c}{STS12} \\
        \cmidrule(lr){2-3} \cmidrule(lr){4-5} \cmidrule(lr){6-7}
        $\alpha$ 
            & F1 & AUROC 
            & F1 & AUROC 
            & F1 & AUROC \\
        \midrule
        0.0          
            & 0.676 & 0.654          
            & 0.694 & 0.739
            & 0.783 & 0.828 \\
        \textbf{0.1}          
            & 0.681 & 0.664 
            & 0.745 & 0.781
            & \textbf{0.786} & \textbf{0.834} \\
        0.2
            & 0.683 & 0.669
            & 0.760 & 0.790
            & 0.780 & 0.834 \\
        \textbf{0.3} 
            & \textbf{0.692} & \textbf{0.673} 
            & 0.764 & 0.793 
            & 0.780 & 0.831 \\
        0.4
            & 0.680 & 0.674
            & 0.764 & 0.795
            & 0.780 & 0.827 \\
        \textbf{0.5}
            & 0.680 & 0.675
            & \textbf{0.766} & \textbf{0.795} 
            & 0.775 & 0.824 \\
        0.6
            & 0.681 & 0.674
            & 0.764 & 0.795
            & 0.768 & 0.820 \\
        0.7
            & 0.688 & 0.674
            & 0.764 & 0.795
            & 0.764 & 0.815 \\
        0.8
            & 0.684 & 0.672
            & 0.764 & 0.795
            & 0.762 & 0.810 \\
        0.9
            & 0.683 & 0.671
            & 0.764 & 0.795
            & 0.756 & 0.805 \\
        1.0
            & 0.681 & 0.671
            & 0.764 & 0.731
            & 0.756 & 0.790 \\
        \bottomrule
    \end{tabular}
\end{table}


\subsection*{B.2 KG-Perturbed Paragraph Datasets}


Table~\ref{tab:alpha_kgpara} presents results for the five KG-perturbed paragraph datasets. Four of the five datasets favour a balanced blend of structural and dense
signal: SK-Codex~400 and SK-Combined peak at $\alpha = 0.5$, SK-GloBI at $\alpha = 0.6$, and SK-Oregano at $\alpha = 0.4$. SK-FindKG is the single exception,
where the pure KG embedding ($\alpha = 0.0$) yields the highest F1 of 0.767, suggesting that the structural signal in FindKG is particularly discriminative and is diluted rather than enhanced by dense representations.

\begin{table}[H]
    \centering
    \caption{S3KG $\alpha$ sweep on KG-perturbed paragraph datasets (F1 / AUROC).
    The best $\alpha$ per dataset by F1 is shown in \textbf{bold}.}
    \label{tab:alpha_kgpara}
    \begin{tabular}{c cc cc cc cc cc}
        \toprule
        {}
            & \multicolumn{2}{c}{SK-Codex~400}
            & \multicolumn{2}{c}{SK-Combined}
            & \multicolumn{2}{c}{SK-FindKG}
            & \multicolumn{2}{c}{SK-GloBI}
            & \multicolumn{2}{c}{SK-Oregano} \\
        \cmidrule(lr){2-3} \cmidrule(lr){4-5} \cmidrule(lr){6-7} \cmidrule(lr){8-9}\cmidrule(lr){10-11}
        $\alpha$ 
            & F1 & AUROC 
            & F1 & AUROC 
            & F1 & AUROC 
            & F1 & AUROC 
            & F1 & AUROC \\
        \midrule
        \textbf{0.0}
            & 0.871 & 0.969
            & 0.776 & 0.801
            & \textbf{0.767} & \textbf{0.796} 
            & 0.811 & 0.898
            & 0.777 & 0.884 \\
        0.1
            & 0.922 & 0.973
            & 0.791 & 0.825
            & 0.760 & 0.798
            & 0.883 & 0.934
            & 0.803 & 0.892 \\
        0.2
            & 0.927 & 0.973
            & 0.819 & 0.829
            & 0.762 & 0.793
            & 0.888 & 0.937
            & 0.811 & 0.892 \\
        0.3          
            & 0.927 & 0.973
            & 0.828 & 0.830
            & 0.748 & 0.788
            & 0.889 & 0.936
            & 0.812 & 0.892 \\
        \textbf{0.4}
            & 0.929 & 0.973
            & 0.833 & 0.829
            & 0.744 & 0.783
            & 0.891 & 0.936
            & \textbf{0.812} & \textbf{0.892} \\
        \textbf{0.5} 
            & \textbf{0.932} & \textbf{0.973} 
            & \textbf{0.834} & \textbf{0.829} 
            & 0.741 & 0.779
            & 0.890 & 0.935
            & 0.810 & 0.892 \\
        \textbf{0.6}
            & 0.932 & 0.973
            & 0.832 & 0.828
            & 0.736 & 0.776
            & \textbf{0.892} & \textbf{0.935} 
            & 0.812 & 0.892 \\
        0.7
            & 0.932 & 0.973
            & 0.833 & 0.828
            & 0.734 & 0.772
            & 0.892 & 0.935
            & 0.812 & 0.891 \\
        0.8
            & 0.932 & 0.973
            & 0.832 & 0.828
            & 0.731 & 0.769
            & 0.892 & 0.935
            & 0.810 & 0.891 \\
        0.9
            & 0.932 & 0.973
            & 0.828 & 0.827
            & 0.729 & 0.765
            & 0.892 & 0.934
            & 0.810 & 0.891 \\
        1.0
            & 0.932 & 0.932
            & 0.828 & 0.799
            & 0.728 & 0.756
            & 0.892 & 0.917
            & 0.810 & 0.770 \\
        \bottomrule
    \end{tabular}
\end{table}


\subsection*{B.3 Wikipedia Entity-Swap (Anti-Circularity Control)}


Table~\ref{tab:alpha_wikiswap} presents the $\alpha$ sweep for the Wikipedia Entity-Swap dataset, which serves as an anti-circularity control. The best performance is achieved at $\alpha = 0.1$ (F1 = 0.872, AUROC = 0.890,
Precision = 0.780), confirming that even a small contribution from the KG structural embedding improves over the pure dense baseline, while heavier structural weighting ($\alpha \geq 0.2$) offers no further benefit.

\begin{table}[H]
    \centering
    \caption{S3KG $\alpha$ sweep on Wikipedia Entity-Swap (F1 / AUROC / Precision / Recall).
    The best $\alpha$ by F1 is shown in \textbf{bold}.}
    \label{tab:alpha_wikiswap}
    \begin{tabular}{c cccc}
        \toprule
        $\alpha$ 
            & F1 & AUROC & Precision & Recall \\
        \midrule
        0.0
            & 0.821 & 0.892 & 0.698 & 0.995 \\
        \textbf{0.1} 
            & \textbf{0.872} & \textbf{0.890} & \textbf{0.780} & 0.990 \\
        0.2
            & 0.869 & 0.890 & 0.783 & 0.975 \\
        0.3
            & 0.868 & 0.890 & 0.773 & 0.990 \\
        0.4
            & 0.865 & 0.890 & 0.777 & 0.975 \\
        0.5
            & 0.868 & 0.890 & 0.773 & 0.990 \\
        0.6
            & 0.865 & 0.889 & 0.777 & 0.975 \\
        0.7
            & 0.865 & 0.889 & 0.777 & 0.975 \\
        0.8
            & 0.865 & 0.889 & 0.777 & 0.975 \\
        0.9
            & 0.865 & 0.889 & 0.777 & 0.975 \\
        1.0 
            & 0.865 & 0.852 & 0.777 & 0.975 \\
        \bottomrule
    \end{tabular}
\end{table}

%
%
\section{Temperature Sensitivity Results}
\label{sec:appendix_temp}


The sampling temperature is a key hyperparameter governing how deterministically a language model generates text. A temperature of zero corresponds to greedy decoding, where the model always selects the most probable next token, yielding highly consistent and context-adherent outputs. As temperature increases, the sampling distribution becomes broader, allowing the model to explore a wider range of responses. Low temperatures such as 0.3 preserve factual grounding while introducing modest lexical variation, whereas a mid-range value of 0.7 is commonly adopted in practice to balance fluency and diversity. At temperature 1.0, the model samples directly from its raw output distribution, producing the most varied responses but with a greater risk of factual drift away from the provided context.

We selected $T \in \{0.0, 0.3, 0.7, 1.0\}$ to span the full practical operating range of instruction-tuned models, from fully deterministic inference to high-entropy generation. This allows us to examine whether contextual faithfulness, as measured by GoldSim and CtxSim, is robust to generation stochasticity or degrades meaningfully as randomness increases. Tables~\ref{tab:temp_gold} and~\ref{tab:temp_ctx} report these results. Most models remain stable within $\pm$0.01--0.02 across all settings; the notable exception is Falcon-7B on MesaQA, where GoldSim falls from 0.6036 at $T{=}0.0$ to 0.4662 at $T{=}1.0$, indicating that higher sampling randomness substantially degrades factual alignment for this model.

\begin{table}[htbp]
    \centering
    \caption{Mean GoldSim per model across temperatures.}
    \label{tab:temp_gold}
    \begin{tabular}{l l cccc}
        \toprule
        \textbf{Dataset} 
            & \textbf{Model} 
            & \textbf{T=0.0} & \textbf{T=0.3} & \textbf{T=0.7} & \textbf{T=1.0} \\
        \midrule
        {}
            & Llama-2-7B 
            & 0.6570 & \textbf{0.6588} & 0.6580 & 0.6406 \\
        MesaQA
            & Gemma-7B   
            & 0.6889 & \textbf{0.7060} & 0.7015 & 0.6931 \\
        {}
            & Mistral-7B 
            & 0.6491 & \textbf{0.6567} & 0.6523 & 0.6444 \\
        {}
            & Falcon-7B  
            & 0.6036 & \textbf{0.6085} & 0.5558 & 0.4662 \\
        \midrule
        {}
            & Llama-2-7B 
            & \textbf{0.5220} & 0.5185 & 0.5143 & 0.5087 \\
        PubMedQA 
            & Gemma-7B   
            & \textbf{0.5235} & 0.5149 & 0.5153 & 0.5204 \\
        {}
            & Mistral-7B 
            & \textbf{0.5138} & 0.5121 & 0.5085 & 0.5023 \\
        {}
            & Falcon-7B  
            & \textbf{0.4541} & 0.4330 & 0.4184 & 0.3852 \\
        \bottomrule
    \end{tabular}
\end{table}

\begin{table}[htbp]
    \centering
    \caption{Mean CtxSim per model across temperatures.}
    \label{tab:temp_ctx}
    \begin{tabular}{l l cccc}
        \toprule
        \textbf{Dataset} 
            & \textbf{Model} 
            & \textbf{T=0.0} & \textbf{T=0.3} & \textbf{T=0.7} & \textbf{T=1.0} \\
        \midrule
        {}
            & Llama-2-7B 
            & 0.7236 & 0.7212 & \textbf{0.7248} & 0.7026 \\
        MesaQA   
            & Gemma-7B   
            & 0.7031 & \textbf{0.7217} & 0.7045 & 0.7013 \\
        {}
            & Mistral-7B 
            & 0.7356 & \textbf{0.7585} & 0.7287 & 0.7223 \\
        {}
            & Falcon-7B  
            & 0.6458 & \textbf{0.6571} & 0.6122 & 0.5195 \\
        \midrule
        {}
            & Llama-2-7B 
            & 0.6567 & 0.6563 & \textbf{0.6594} & 0.6468 \\
        PubMedQA 
            & Gemma-7B   
            & 0.6401 & \textbf{0.6540} & 0.6434 & 0.6319 \\
        {}
            & Mistral-7B 
            & 0.7331 & \textbf{0.7352} & 0.7314 & 0.7003 \\
        {}
            & Falcon-7B  
            & \textbf{0.5560} & 0.5165 & 0.5120 & 0.4571 \\
        \bottomrule
    \end{tabular}
\end{table}

%
\label{sec:appendix_benchmark}


This appendix provides complete benchmark results referenced in the main paper, including short-text datasets (Section~\ref{sec:benchmark}), KG-perturbed paragraph datasets, the Wikipedia Entity-Swap anti-circularity control, and a summary heatmap visualization.


\subsection*{D.1 Short-Text Datasets}


Table~\ref{tab:short} presents the full F1 and AUROC results for MRPC, PAWS-Wiki, and STS12. S3KG achieves competitive performance, with sentence-T5-base leading on STS12 due to the short-text nature of the dataset.

\begin{table}[htbp]
    \centering
    \caption{Performance on short-text datasets (F1 / AUROC).}
    \label{tab:short}
    \setlength{\tabcolsep}{4pt}
    \begin{tabular}{l ccc}
        \toprule
        \textbf{Method} 
            & \textbf{MRPC} & \textbf{PAWS-Wiki} & \textbf{STS12} \\
        \midrule
        S3KG
            & 0.692 / 0.673 & 0.766 / 0.795 & 0.786 / 0.834 \\
        \midrule
        ROUGE-1
            & 0.745 / 0.784 & 0.678 / 0.490 & 0.725 / 0.754 \\
        ROUGE-2
            & 0.720 / 0.721 & 0.715 / 0.721 & 0.681 / 0.656 \\
        ROUGE-L
            & 0.729 / 0.760 & 0.735 / 0.807 & 0.703 / 0.710 \\
        BLEU
            & 0.687 / 0.677 & 0.716 / 0.747 & 0.671 / 0.644 \\
        BERTScore
            & 0.758 / 0.816 & 0.691 / 0.702 & 0.682 / 0.636 \\
        MiniLM
            & 0.723 / 0.748 & 0.687 / 0.638 & 0.833 / 0.894 \\
        sentence-T5-base 
            & \textbf{0.766} / \textbf{0.816} & 0.674 / 0.668 & \textbf{0.853} / \textbf{0.928} \\
        \bottomrule
    \end{tabular}
\end{table}


\subsection*{D.2 KG-Perturbed Paragraph Datasets}


Table~\ref{tab:kg} reports results for the five KG-perturbed paragraph datasets. S3KG achieves the highest F1 on four of five datasets, with gains up to $+7.6$ F1 points on SK-GloBI.

\begin{table}[htbp]
    \centering
    \caption{Performance on KG-perturbed paragraph datasets (F1 / AUROC).}
    \label{tab:kg}
    \setlength{\tabcolsep}{2.8pt}
    \renewcommand{\arraystretch}{1.05}
    \begin{tabular}{l ccccc}
        \toprule
        \textbf{Method} 
            & \textbf{C400} & \textbf{Comb.} & \textbf{Find} & \textbf{GloBI} & \textbf{Oreg.} \\
        \midrule
        S3KG      
            & \textbf{0.932}/\textbf{0.973} & \textbf{0.834}/\textbf{0.829} & 0.767/0.796 & \textbf{0.892}/\textbf{0.935} & \textbf{0.812}/\textbf{0.892} \\
        \midrule
        ROUGE-1   
            & 0.835/0.917 & 0.732/0.728 & 0.745/0.745 & 0.784/0.833 & 0.745/0.782 \\
        ROUGE-2
            & 0.822/0.894 & 0.707/0.711 & 0.717/0.706 & 0.776/0.833 & 0.752/0.791 \\
        ROUGE-L
            & 0.792/0.855 & 0.722/0.717 & 0.719/0.721 & 0.763/0.800 & 0.792/0.835 \\
        BLEU      
            & 0.806/0.884 & 0.715/0.708 & 0.711/0.710 & 0.775/0.819 & 0.745/0.794 \\
        BERTScore 
            & 0.823/0.916 & 0.757/0.792 & 0.739/0.761 & 0.816/0.871 & 0.743/0.798 \\
        MiniLM    
            & 0.875/0.943 & 0.770/0.789 & 0.802/0.844 & 0.780/0.817 & 0.773/0.814 \\
        sent-T5   
            & 0.876/0.944 & 0.770/0.827 & \textbf{0.848}/\textbf{0.902} & 0.728/0.760 & 0.797/0.871 \\
        \bottomrule
        \multicolumn{6}{l}{%
        \footnotesize C400: SK-Codex 400; Comb.: SK-Combined; 
        Find: SK-FindKG; Oreg.: SK-Oregano.}
    \end{tabular}
\end{table}


\subsection*{D.3 Wikipedia Entity-Swap Results}


Table~\ref{tab:swap} presents results on the Wikipedia Entity-Swap anti-circularity control. S3KG achieves the highest F1 (0.872) and AUC (0.890), confirming that gains are not an artefact of the KG construction pipeline.\footnote{ROUGE-1 is excluded due to a surface-form artefact: entity-swapped pairs share nearly all surrounding tokens, making unigram overlap trivially near-perfect.}

\begin{table}[htbp]
    \centering
    \caption{Results on Wikipedia Entity-Swap ($N=400$).}
    \label{tab:swap}
    \begin{tabular}{l cccc}
        \toprule
        \textbf{Method} 
            & \textbf{F1} & \textbf{AUC} & \textbf{Prec.} & \textbf{Rec.} \\
        \midrule
        S3KG
            & \textbf{0.872} & \textbf{0.890} & \textbf{0.780} & 0.990 \\
        \midrule
        ROUGE-2
            & 0.860 & 0.772 & 0.773 & 0.970 \\
        ROUGE-L
            & 0.729 & 0.311 & 0.573 & \textbf{1.000} \\
        BLEU
            & 0.868 & 0.745 & 0.776 & 0.985 \\
        BERTScore
            & 0.747 & 0.645 & 0.615 & 0.950 \\
        MiniLM
            & 0.821 & 0.811 & 0.729 & 0.940 \\
        sentence-T5-base 
            & 0.762 & 0.806 & 0.621 & 0.985 \\
        \bottomrule
    \end{tabular}
\end{table}


\subsection*{D.4 Best S3KG Variant Summary}


Table~\ref{tab:summary} summarizes the best-performing $\alpha$ variant per dataset, selected by maximum F1 via grid search over $\alpha \in \{0.0, 0.1, \ldots, 1.0\}$.

\begin{table}[htbp]
    \centering
    \caption{Best S3KG variant per dataset.}
    \label{tab:summary}
    \setlength{\tabcolsep}{3pt}
    \begin{tabular}{l c c}
        \toprule
        \textbf{Dataset} 
            & \textbf{F1/AUROC} 
            & \textbf{Verdict} \\
        \midrule
        MRPC ($\alpha = 0.3$)        
            & 0.692/0.673 
            & Behind (sparse KG) \\
        PAWS-Wiki ($\alpha = 0.5$)   
            & 0.766/0.795 
            & Strong ($+$3.1 F1) \\
        STS12 ($\alpha = 0.1$)
            & 0.786/0.834 
            & Behind (short text) \\
        SK-Codex 400 ($\alpha = 0.5$) 
            & 0.932/0.973 
            & Strong ($+$5.7 F1) \\
        SK-Combined ($\alpha = 0.5$) 
            & 0.834/0.829 
            & Strong ($+$6.0 F1) \\
        SK-FindKG ($\alpha = 0.0$)   
            & 0.767/0.796 
            & Behind (domain noise) \\
        SK-GloBI ($\alpha = 0.6$)
            & 0.892/0.935 
            & Strong ($+$7.6 F1) \\
        SK-Oregano ($\alpha = 0.4$)  
            & 0.812/0.892 
            & Comparable ($+$1.2) \\
        Wiki Swap ($\alpha = 0.1$)   
            & 0.872/0.890 
            & Best meaningful score \\
        \bottomrule
    \end{tabular}
\end{table}


\subsection*{D.5 Performance Heatmap}

\begin{figure}[htbp]
    \centering
    \includegraphics[width=\linewidth]{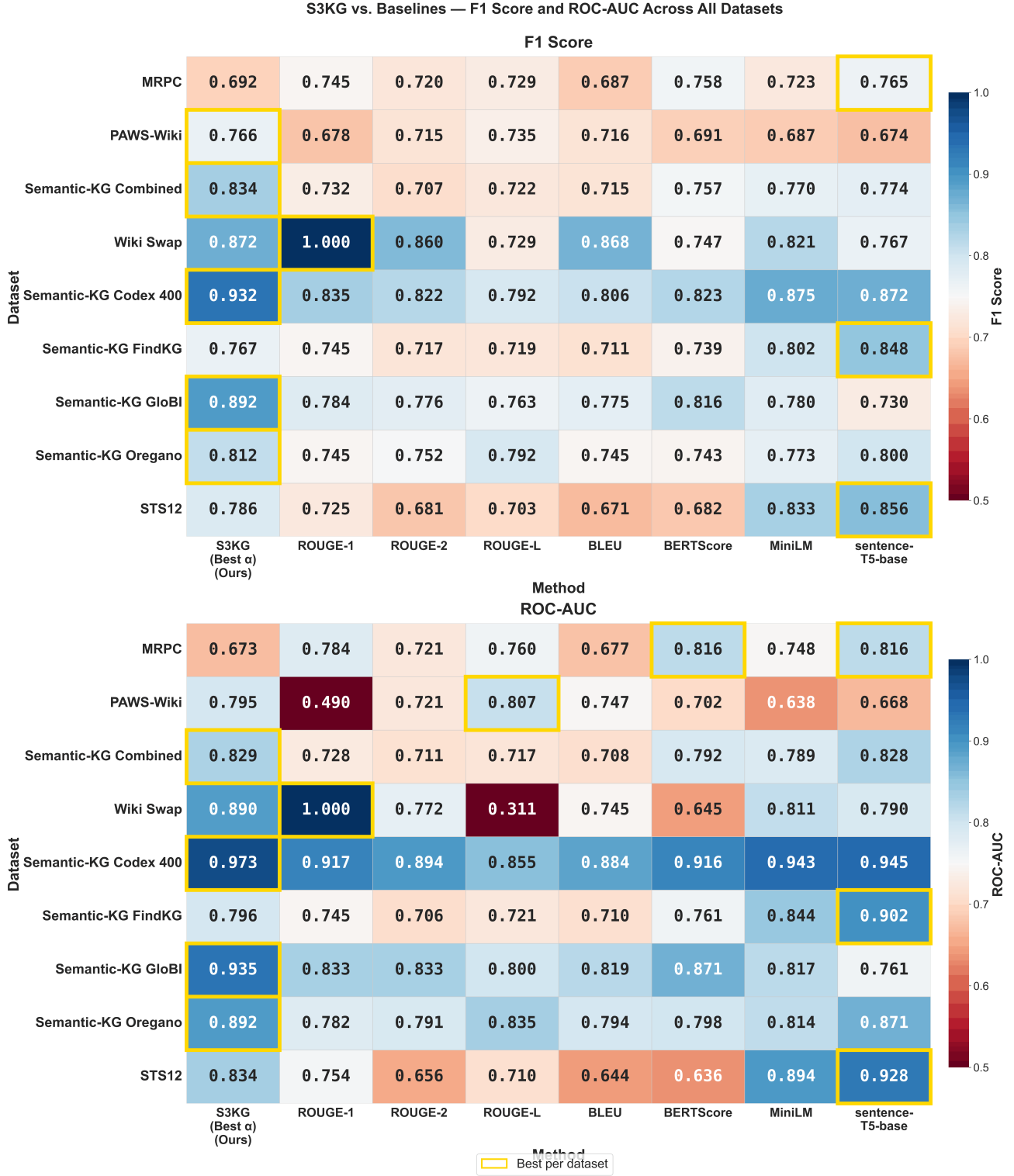}
    \caption{F1 score (left) and AUROC (right) of S3KG and baseline methods across all benchmark datasets. Gold borders indicate the best-performing method per dataset. S3KG (navy border, leftmost column) represents the best-performing $\alpha$ variant selected per dataset from the full $\alpha$ sweep (Appendix~\ref{sec:appendix_alpha}).}
    \label{fig:heatmap}
\end{figure}

Figure~\ref{fig:heatmap} visualizes the F1 and AUROC scores of S3KG and all baseline methods across all benchmark datasets. Gold borders indicate the best-performing method per dataset.

%
\label{sec:appendix_worked}


This appendix provides a detailed worked example from the PAWS-Wiki dataset, illustrating how S3KG correctly identifies semantic opposition where ROUGE-1 fails. Table~\ref{tab:worked_example} compares a positive (similar) pair and a negative (not similar) pair.

\begin{table}[htbp]
    \centering
    \caption{S3KG pipeline scores for two PAWS-Wiki sentence pairs.}
    \label{tab:worked_example}
    \setlength{\tabcolsep}{5pt}
    \begin{tabular}{p{2.2cm} p{6.5cm} p{6.5cm}}
        \toprule
        {}
            & \textbf{Positive Example} 
            & \textbf{Negative Example} \\
        \midrule
        \textbf{Text $s_1$} 
            & \textit{His father returned as a finished violinist of the Russian School to Bombay.} 
            & \textit{Renzo Furlan won 6--3, 6--4 against Thomas Johansson in the finals.} \\[4pt]
        \textbf{Text $s_2$} 
            & \textit{His father returned to Bombay as a finished violinist of the Russian school.} 
            & \textit{Thomas Johansson won 6--3, 6--4 against Renzo Furlan in the finals.} \\[4pt]
        \textbf{Triplet ($s_1$)} 
            & \small{(father, returned\_to, Bombay)} 
            & \small{(Renzo Furlan, won\_against, Thomas Johansson)} \\[2pt]
        \textbf{Triplet ($s_2$)} 
            & \small{(father, returned\_to, Bombay)} 
            & \small{(Thomas Johansson, won\_against, Renzo Furlan)} \\[4pt]
        \midrule
        $\text{sim}_\text{TP}$ ($\alpha{=}0$)  
            & 1.0000 & 0.8310 \\
        $\text{sim}_\text{ST}$ ($\alpha{=}1$)  
            & 1.0000 
            & 0.1361 \\
        $\text{sim}_\alpha$ ($\alpha{=}0.5$)   
            & \textbf{1.0000} 
            & \textbf{0.4835} \\
        Threshold $\tau$                        
            & 0.92 
            & 0.92 \\
        \midrule
        S3KG prediction   
            & Similar \checkmark 
            & Not Similar \checkmark \\
        ROUGE-1 score
            & 1.0000
            & 1.0000 \\
        ROUGE-1 prediction 
            & Similar \checkmark 
            & Similar $\times$ \\
        \midrule
        \textbf{True label} & Positive (1) & Negative (0) \\
        \bottomrule
    \end{tabular}
\end{table}

In the positive example, word-order variation yields identical KG triplets and $\text{sim}_\alpha = 1.0$, correctly predicting \textit{Similar}. In the negative example, reversed subject--object roles expose semantic opposition despite identical surface tokens: ROUGE-1 incorrectly predicts \textit{Similar}, while S3KG correctly predicts \textit{Not Similar}.


\end{document}